\documentclass[runningheads]{llncs}
\usepackage[T1]{fontenc}
\usepackage{graphicx}
\usepackage{xcolor}
\usepackage{amsmath}
\usepackage{amssymb}
\usepackage{multirow}
\usepackage{booktabs}

\DeclareMathOperator*{\argmin}{arg\,min}
\begin{document}
\title{On the Viability of Semi-Supervised Segmentation Methods for Statistical Shape Modeling}
\titlerunning{Semi-Supervised Segmentation Methods for SSM}
%
\author{Asma Khan$^*$ \and
Tushar Kataria$^*$  \and
Janmesh Ukey \and
Shireen Elhabian }
\authorrunning{A. Khan et al.}
%
\institute{Kahlert School of Computing, University of Utah, Salt Lake City, USA \and
Scientific Computing and Imaging Institute, University of Utah, Salt Lake City, USA \\
\email{  (asmak,tushar.kataria,janmesh,shireen)@sci.utah.edu}\\
* Equal Contribution}
\maketitle              
\begin{abstract}
 Statistical Shape Models (SSMs) excel at identifying population level anatomical variations, which is at the core of various clinical and biomedical applications, including morphology-based diagnostics and surgical planning.
However, the effectiveness of SSMs is often constrained by the necessity for expert-driven manual segmentation, a time-intensive and expensive process that restricts their broader utility. 
%
While deep learning approaches can estimate SSMs directly from unsegmented images, they still require segmentation annotations for SSM training. Although segmentations are unnecessary during deployment, the challenge of acquiring training manual annotations remains.
%
Semi-supervised anatomical segmentation offers a promising solution to the annotation burden by leveraging a small labelled dataset alongside abundant unlabelled images. However, its effectiveness for downstream statistical shape model (SSM) construction remains largely unexplored. In this study, we systematically evaluate semi-supervised methods as alternatives to manual segmentation under low-annotation settings, using their predicted segmentations to construct SSMs and establish a comprehensive performance benchmark.
Our findings reveal a clear divide in performance: while certain methods yield noisy segmentations that degrade SSM quality, others accurately capture the population's modes of variation comparable to those obtained from manual-segmentation SSMs despite a 60–80\% reduction in manual annotation requirements.

\end{abstract}
\section{Introduction}
Statistical shape models (SSMs) provide a quantitative framework for analyzing anatomical variations across populations, enabling the identification of normative trends and deviations and facilitating the development of diagnostic tools and surgical planning systems \cite{goparaju2022benchmarking}.
Constructing SSMs is contingent upon the accurate segmentation of the target anatomy. This process is both time-consuming and resource-intensive, often hindered by the scarcity of medical expertise necessary for precise segmentation.
Recent advances in deep learning have facilitated the direct estimation of SSMs from unsegmented images, thereby bypassing the need for segmentation during inference 
\cite{bhalodia2018deepssm,adams2020uncertain,adams2022images,tao2022learning,milletari2017integrating,xie2016deepshape,raju2022deep,karanam2023adassm,bhalodia2024deepssm,ukey2023localization,ukey2024massm,bhalodia2024deepssm,xu2023image2ssm,iyer2023mesh2ssm,iyer2024scorp,adams2024weakly}. 
However, these deep learning methods still necessitate anatomy segmentation to construct SSMs for training.

Automated deep learning-based anatomical segmentation can reduce the burden of generating segmentations for statistical shape model (SSM) construction. However, conventional supervised segmentation methods still require large amounts of manually annotated data for training. To address this limitation, numerous semi-supervised approaches have been proposed \cite{goparaju2022benchmarking,luo2021semi,bai2023bidirectional,yu2019uncertainty,tarvainen2017mean}. These methods leverage a small set of labelled volumes alongside a larger unlabelled dataset and employ techniques such as data augmentation \cite{bai2023bidirectional}, pseudo-labeling \cite{bai2023bidirectional,tarvainen2017mean,yu2019uncertainty}, consistency regularization \cite{tarvainen2017mean,wu2021r}, and entropy minimization \cite{vu2019advent} to achieve high-quality segmentations.
Several methods (e.g., \cite{li2022pln,yeung2021sli2vol,nihalaani2024estimation}) further exploit the inherent 3D structure of medical images, reducing annotation requirements from full-volume segmentations to a small number of annotated slices. 

\begin{figure}[!thb]
    \centering
    \includegraphics[width=1.0\linewidth]{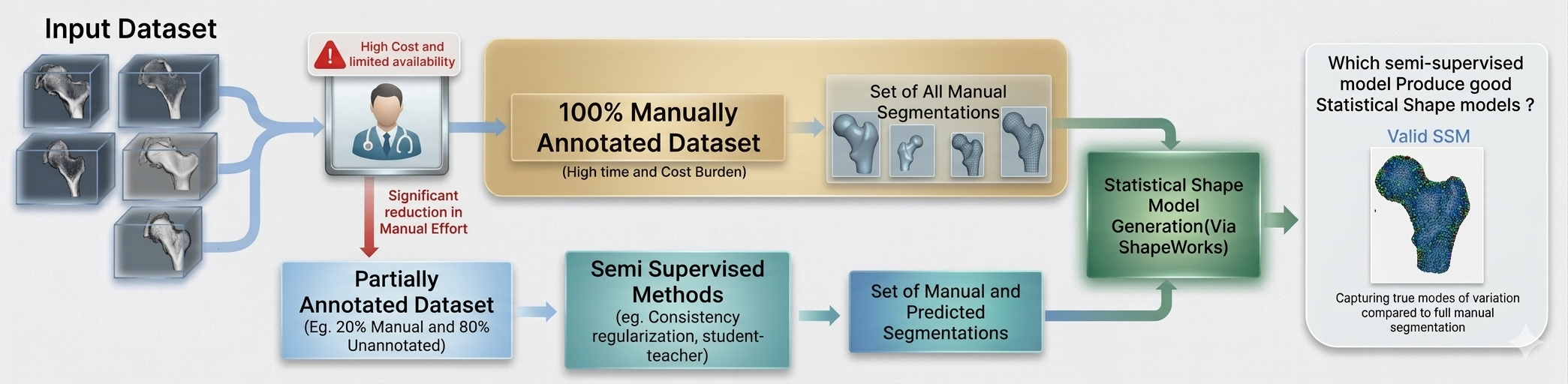}
    \vspace{-1em}
    \caption{\textbf{Benchmarking pipeline.} Semi-supervised segmentation models trained on limited labelled data and unlabelled images generate anatomical segmentations for SSM construction, which are benchmarked against SSMs from manual segmentations to assess the impact of reduced annotation on shape modeling.}
    \label{fig:main_figure}
    \vspace{-2em}
\end{figure}

Despite the growing number of annotation-efficient segmentation methods, there are no established guidelines for assessing their suitability as alternatives to manual segmentation for statistical shape model (SSM) construction. Existing evaluations focus primarily on segmentation accuracy, which does not necessarily reflect the quality of the resulting SSMs. Predicted segmentations may achieve high overlap scores while altering anatomical boundaries and the statistical modes of shape variation. Therefore, evaluating these methods requires assessing the quality of the downstream SSMs rather than segmentation accuracy alone. To our knowledge, no comparable benchmark exists, making our evaluation a useful reference for developing medical SSMs with limited annotation (shown in Figure ~\ref{fig:main_figure}).

In this paper, we present the first benchmark for evaluating whether semi-supervised segmentation methods can serve as annotation-efficient alternatives to manual segmentation for statistical shape model (SSM) construction. Our benchmark compares SSMs generated from semi-supervised segmentations with those built from manual annotations, assessing their suitability for downstream population-level shape analysis under limited annotation settings. By identifying methods that preserve SSM quality while reducing annotation requirements, this work aims to improve the accessibility and practical adoption of SSMs in medical applications. The main contributions of this manuscript are:
\begin{itemize} 
\item We establish the first benchmark for evaluating semi-supervised segmentation methods based on the quality of the statistical shape models they produce, using manual-segmentation-based SSMs as the reference.
\item We evaluate SSM quality across varying annotation settings, including different proportions (20\% and 40\%) and annotation types (full-volume and sparse-slice) used to train semi-supervised segmentation models.
\item We provide a comprehensive quantitative and qualitative analysis of downstream SSM performance, identifying the strengths and failure modes of current semi-supervised segmentation methods. 
\end{itemize}

\section{Methods}
To determine whether semi-supervised methods can effectively replace manual annotations for SSM, we seek to address the following key questions:-
\begin{itemize}
    \item \textit{Do SSM models created using semi-supervised model predictions capture the same modes of variation compared to an SSM model using all manual segmentations?}
    \item \textit{Are some semi-supervised methods more reliable than others?}
    \item \textit{Does increasing the amount of annotated data for semi-supervised models improve SSM for the population cohort?}
\end{itemize}

\textbf{Selected Semi-Supervised Methods.} {Semi-supervised medical image segmentation leverages partially labeled and all unlabeled data during training. We selected eight distinct methods that utilize labeled and unlabeled data in various ways, categorizing them into three distinct groups: (A) \textit{Student-teacher methods}:- {Mean Teacher (MT)\cite{tarvainen2017mean}}  \& {Uncertainty-Aware Mean Teacher (UA-MT)\cite{yu2019uncertainty}}, (B) \textit{Pseudo-labeling methods:-} {Mutual Correction Framework (MCF) \cite{wang2023mcf}}, {Orthogonal Annotation (DeSCO) \cite{cai2023orthogonal}}, {Bidirectional Copy-Paste (BCP)\cite{bai2023bidirectional}} and {Correlation-Aware Mutual Learning (CAML) \cite{gao2023correlation}}, and (C)\textit{ Multi-task learning methods:-} {Dual-task Consistency (DTC) \cite{cai2023orthogonal}}, and {Shape-Aware Semi-supervised (SASSnet) \cite{li2020shape}}.} A brief description of each of these methods is added below:

{
\textbf{Mean Teacher (MT)\cite{tarvainen2017mean}.} Mean Teacher uses consistency regularization between  \textit{student}-\textit{teacher} training paradigm. The \textit{student} model is trained using the supervised loss on the labeled set, whereas the \textit{teacher}'s parameters are updated as an exponential moving average of the student model. Consistency loss on unlabeled data aligns the prediction between the two models.}

{\textbf{Uncertainty-Aware Mean Teacher (UA-MT)\cite{yu2019uncertainty}.} An extension of MT where the  \textit{teacher} model also estimates the uncertainty of each target prediction with Monte Carlo sampling. The uncertainty is used to preserve only the reliable predictions when calculating the consistency loss.}

{\textbf{Mutual Correction Framework (MCF) \cite{wang2023mcf}}. MCF employs a student-teacher model paradigm to address cognitive biases learned during training, applying a rectification loss in regions where the model frequently makes errors. It then uses the dice scores of segmentations predicted by both models to determine the pseudo-labels for unlabeled data.}

{\textbf{Bidirectional Copy-Paste (BCP)\cite{bai2023bidirectional}.} BCP integrates a bidirectional copy-paste framework into the Mean Teacher architecture. The student network receives input created by pasting random crops from labeled images onto unlabeled images and vice versa (bi-direction copy). The supervision of the \textit{student} network involves combining both ground-truth labels and pseudo-labels generated by the \textit{teacher} network using the same bidirectional copy-paste process.}



{\textbf{Correlation-Aware Mutual Learning (CAML) \cite{gao2023correlation}}. CAML introduces two modules: Cross-sample Mutual Attention, which uses transformers to compute attention between labeled and unlabeled samples within the same batch, and Omni-Correlation Consistency, a contrastive loss based on pixel-level latent space projections between labeled and unlabeled volumes. These modules enhance the transfer of information between labeled and unlabeled data, leading to improved segmentation performance.}

{\textbf{Orthogonal Annotation (DeSCO) \cite{cai2023orthogonal}}. This paper presents a novel orthogonal annotation strategy for semi-supervised 3D medical image segmentation, where only three orthogonal slices are labeled and the remaining volumes are unlabeled. It employs cross-consistency and pseudo-labeling to enhance performance. This method significantly reduces the annotation effort required, making it more feasible for broader deployment.}

{\textbf{Dual-task Consistency (DTC) \cite{luo2021semi}}. Instead of applying consistency loss between labeled and unlabeled data, as seen in MT \cite{tarvainen2017mean}, UA-MT \cite{yu2019uncertainty}, and similar methods \cite{wu2021r}, DTC introduces a multi-task network to enforce consistency between two tasks. The chosen tasks are segmentation and level-set prediction, where the level-set function's zero level corresponds to the contour of the segmented anatomy. The model is trained to ensure consistent predictions across both tasks for both labeled and unlabeled data.}

{\textbf{Shape-Aware Semi-supervised (SASSnet) \cite{li2020shape}}. SASSnet introduces a shape-aware semi-supervised strategy that leverages multitasking to predict both segmentation and the signed distance map (SDM) simultaneously. Additionally, it employs an adversarial loss between the SDM predictions for labeled and unlabeled data, which encourages the model to learn shape-aware features.}

\subsection{Experimental Design}
To offer a thorough evaluation of semi-supervised methods for constructing Statistical Shape Models (SSM), we design two experimental strategies that enable detailed analysis and support the future deployment of these methods for end-user applications.

{Consider a dataset $\mathcal{D}$, which consists of a collection of anatomical images ($\mathcal{I}$) along with their corresponding manual segmentations ($\mathcal{S}$). The dataset is split into training $\mathcal{D}_{Train}$ and testing set $\mathcal{D}_{test}$, such that  $\mathcal{D}_{Train} \cap \mathcal{D}_{Test} = \emptyset$ and $\mathcal{D}_{Train} \cup \mathcal{D}_{Test} = \mathcal{D}$. 

We define the SSM representation of an anatomy as $\Phi$. The SSM is formulated as an optimization process aimed at minimizing the entropy of particles (or point clouds) on the surface of the anatomy \cite{cates2017shapeworks}. This process is applied to all segmentations in the training dataset, denoted as $\mathcal{F}_{\Phi}(\cdot)$, where the function parameters consist of the segmentations and the initial particle locations.}
${\Phi}_{Train}$ establishes the best training SSM representation, as it utilizes all the manual annotations in the training set to construct an SSM, as described by the equation:
\begin{equation}
\Phi_{\mathrm{train}} = \argmin_{\Phi} \sum_{i \in D_{\mathrm{train}}} F_{\Phi}(S_i; \phi_0), 
\end{equation}

{\noindent where $\phi_0$ represents the initial condition. $\phi_0$ is set to origin for all particles. To generate the test SSM, we apply the same optimization over the test segmentations, with the  ${\Phi}_{Train}$ as initial condition (fixed domain initialization \cite{bhalodia2024deepssm}),}

\begin{equation}
\Phi_{\mathrm{test}} = \argmin_{\Phi} \sum_{i \in D_{\mathrm{test}}} F_{\Phi}(S_i; \Phi_{\mathrm{train}}).
\end{equation}

{${\Phi}_{Test}$ represents the manual-segmentation reference SSM on the test set, reflecting the optimal SSM representation achievable using all manual annotations. However, since these test samples are unseen by the SSM model, distribution shifts between the training and test segmentations—such as variations in size and contour—can result in generalization errors.}

Let us denote all selected semi-supervised methods discussed above as a set called ${SEMI} \in \{ \text{\textit{MT,UA-MT,BCP,CAML,DeSCO,DTC,MCF,SASSnet}} \}$. The predicted segmentations from these methods can be represented by $\tilde{\mathcal{S}}^{k} \forall k \in {SEMI}$. To evaluate the potential of semi-supervised methods as an alternative to manual segmentations in SSM construction, we investigate two strategies:
\begin{itemize}
    \item \textbf{${\Phi}_{Train}$ available (Strategy 1)}: Instead of using manual segmentations as in equation 2, predicted segmentations from each semi-supervised method ($k \in {SEMI}$) on the test set are used to obtain the SSM representation as, 
    
    \begin{equation}
\widetilde{\Phi}^{k}_{\mathrm{test}}
=
\argmin_{\Phi}
\sum_{i \in D_{\mathrm{test}}}
F_{\Phi}(\widetilde{S}^{k}_i; \Phi_{\mathrm{train}}),
\qquad k \in \mathrm{SEMI}.
\end{equation}
    
compared to ${\Phi}_{Test}$, $ \tilde{{\Phi}}^{k}_{Test}$ includes errors from both distribution shifts between the training and test segmentations and the absence of manual annotations for test set. 
    \item \textbf{$\mathbf{\Phi}_{Train}$ not available (Strategy 2)}: This strategy further limits access to manual segmentations by assuming that only limited training annotations are available for SSM representation on the training set. Only a small amount of segmentation data (20\% or 40\% of manual segmentations) are provided to train the semi-supervised methods, which then generate predicted segmentations for both the training and test sets. For each semi-supervised method ($k \in {SEMI}$) we obtain the SSM representation of the training set as:
    \begin{equation}
    \widetilde{\Phi}^{k}_{\mathrm{train}}
    =
    \argmin_{\Phi}
    \sum_{i \in D_{\mathrm{train}}}
    F_{\Phi}(\widetilde{S}^{k}_i; \phi_0),
    \qquad k \in \mathrm{SEMI}.
    \end{equation}
    where $\phi_0$ is the initial condition. Using $\tilde{{\Phi}}^{k}_{Train}$ representation, we create test SSM as,


\begin{equation}
\widetilde{\Phi}^{k}_{\mathrm{test}}
=
\argmin_{\Phi}
\sum_{i \in D_{\mathrm{test}}}
F_{\Phi}(\widetilde{S}^{k}_i; \widetilde{\Phi}^{k}_{\mathrm{train}}),
\qquad k \in \mathrm{SEMI}.
\end{equation}

compared to ${\Phi}_{Test}$, the $\tilde{{\Phi}}^{k}_{Test}$ representation is impacted by the same errors from Strategy 1, with the added impact due to absence of manual segmentations on the training set. The two strategies help isolate different sources of error: Strategy 1 isolates the generalization error only, whereas Strategy 2 captures both the generalization error and the impact of the unavailability of manual segmentations for constructing the training SSM and testing SSM. 
\end{itemize}


\subsection{Datasets and Metrics for Performance Evaluation}
\textbf{Datasets.} We evaluated our methods on two datasets: the public NAMIC Left Atrium(LA) dataset \cite{bhalodia2024deepssm} and an in-house FEMUR dataset. To assess semi-supervised performance, we tested two levels of labeled data availability (20\% and 40\%) during training. The FEMUR dataset (49 volumes) was split into 40 training samples (20\%: 8 labeled/32 unlabeled; 40\%: 16 labeled/24 unlabeled) and 9 testing samples. Similarly, the NAMIC dataset was split into 50 training samples (20\%: 10 labeled/40 unlabeled; 40\%: 20 labeled/30 unlabeled) and 9 testing samples. We selected the 20\% setting to evaluate a highly annotation-constrained regime using no more than 10 manually labeled training volumes, and the 40\% setting to test whether additional labeled data improves both semi-supervised segmentation performance and downstream SSM quality.

\textbf{Implementation Details.} We used the original codebases with default parameters for all semi-supervised methods. Models were trained on a 12GB NVIDIA 1080 Ti GPU, except for CAML, which required a 40GB NVIDIA A100 due to memory constraints. Finally, ShapeWorks \cite{cates2017shapeworks} was utilized to generate all shape models for analysis given its established efficacy \cite{goparaju2022benchmarking}.

\textbf{Evaluation Metrics.} We evaluate segmentation accuracy using Dice and Jaccard scores, Average Surface Distance (ASD), and the 95th-percentile Hausdorff distance \cite{tarvainen2017mean,yu2019uncertainty,kataria2023pretrain,wang2023mcf,kataria2024unsupervised}. We report HD95 instead of the maximum Hausdorff distance to reduce sensitivity to isolated outlier voxels while preserving sensitivity to boundary errors.

In an ideal statistical shape model (SSM), point locations and neighbourhood relationships are preserved across the segmentations of the anatomy of interest for all patients. This enables meaningful analysis of shape variation across the population. In Statistical Shape Modeling (SSM), aligned corresponding points from all shapes are averaged to create a mean shape representing the typical anatomy. Principal Component Analysis (PCA) is then applied to identify the main modes of shape variation, with each mode describing an independent pattern of anatomical change. These modes can be used to compare healthy and diseased populations, test medical hypotheses, and visualize shape differences by deforming the mean shape along each principal component. Therefore to compare SSM from manual segmentation and predicted segmentation from semi-supervised methods, we compute four metrics based on the PCA projections of SSM particles \cite{cates2017shapeworks,adams2024weakly,bhalodia2024deepssm,adams2024point2ssm}:
\begin{itemize}
\item \textbf{Compactness:} Measured using the area under the curve (AUC). Higher AUC values indicate a more efficient SSM, requiring fewer principal component analysis (PCA) modes (parameters) to represent a given amount of shape variation in the training set.
\item \textbf{Specificity:} The specificity metric measures the extent to which an SSM generates anatomically plausible instances that belong to the shape class represented by the training set. It is computed as the average distance between shapes randomly generated by the SSM and their closest corresponding shapes in the training set.
\item \textbf{Generalization:} Assesses transferability to unseen subjects using the distance between test particle correspondences and their PCA reconstructions (lower is better).
\item \textbf{Grassmannian distance \cite{lim2021grassmannian}:} We introduce this as a new evaluation method to measure the distance between two PCA subspaces for each mode (lower is better).
\end{itemize}

Statistical Shape Models (SSMs) are also assessed qualitatively by comparing their first two modes of variation.
\section{Results and Discussion}

\quad \quad \textbf{Segmentation Performance of Semi-Supervised methods.} We report the segmentation performance on test data in Tables \ref{tab:left_atrium_eval} and \ref{tab:femur_eval}, evaluating each semi-supervised method using 20\% and 40\% labeled training splits. The reported standard deviations quantify inter-subject variability on the held-out test set. Several noteworthy observations emerge from these results:

(1) \textit{Scaling with Annotated Data}: Increasing the amount of annotated training data does not universally guarantee improved performance. While methods like MCF scale exceptionally well, DeSCO and MT exhibit a slight degradation in segmentation accuracy when transitioning from 20\% to 40\% labeled data. DeSCO, in particular, shows severe boundary instability on the LA dataset at 40\% data, where its HD95 more than doubles.

(2) \textit{Methodological Superiority}: MCF emerges as the strongest overall model, particularly at the 40\% data split—closely followed by BCP, which excels at the 20\% split. Conversely, purely Student-Teacher architectures like MT and UA-MT consistently rank among the lowest-performing methods. Overall, architectures utilizing pseudo-labeling (MCF, BCP, CAML) tend to outperform other paradigms on these tasks.

\begin{table}[!htb]
\vspace{-1em}
    \caption{\textbf{Left Atrium Semi-Supervised Segmentation Results.} Metrics include Dice, Jaccard (Jacc.), ASD, and HD95 on test data, reported as mean$_{\text{standard deviation}}$ across nine held-out cases. $\uparrow$/$\downarrow$ indicate higher/lower is better, and the best method per configuration is bolded.}
    \centering
    \setlength{\tabcolsep}{1pt}
    \scalebox{0.86}{
    \begin{tabular}{l|cccc|cccc}
    \toprule
    \multirow{2}{*}{\textbf{Method}} & \multicolumn{4}{c|}{\textbf{20\% Data (10 labeled)}} & \multicolumn{4}{c}{\textbf{40\% Data (20 labeled)}} \\
    \cmidrule(lr){2-5} \cmidrule(lr){6-9}
     & \textbf{Dice} ($\uparrow$) & \textbf{Jacc.} ($\uparrow$) & \textbf{ASD} ($\downarrow$) & \textbf{HD95} ($\downarrow$) & \textbf{Dice} ($\uparrow$) & \textbf{Jacc.} ($\uparrow$) & \textbf{ASD} ($\downarrow$) & \textbf{HD95} ($\downarrow$) \\
    \midrule
    \textbf{DeSCO} & 0.862 $_{0.022}$ & 0.757 $_{0.033}$ & 1.81 $_{0.54}$ & 7.17 $_{2.18}$ & 0.841 $_{0.047}$ & 0.728 $_{0.069}$ & 4.51 $_{1.74}$ & 15.54 $_{6.31}$ \\
    \textbf{SASS}  & 0.855 $_{0.029}$ & 0.748 $_{0.045}$ & 2.16 $_{0.83}$& 8.92 $_{3.12}$& 0.886 $_{0.022}$& 0.795 $_{0.035}$ & 1.42 $_{0.37}$ & 5.61 $_{1.18}$\\
    \textbf{DTC}   & 0.861 $_{0.030}$ & 0.757 $_{0.046}$ & 1.83 $_{0.77}$& 7.80 $_{2.74}$ & 0.881 $_{0.026}$ & 0.789 $_{0.042}$ & 1.49 $_{0.52}$ & 6.19 $_{2.71}$\\
    \textbf{MCF}   & 0.868 $_{0.028}$ & 0.768 $_{0.045}$& 2.01 $_{0.85}$ & 7.78 $_{3.82}$ & \textbf{0.902} $_{0.016}$ & \textbf{0.823} $_{0.027}$ & 1.34 $_{0.33}$& \textbf{4.62} $_{1.70}$\\
    \textbf{CAML}  & 0.889 $_{0.016}$ & 0.800 $_{0.025}$ & \textbf{1.32} $_{0.27}$ & 6.18 $_{1.76}$ & 0.893 $_{0.018}$ & 0.807 $_{0.030}$ & \textbf{1.30} $_{0.35}$ & 6.12 $_{1.56}$ \\
    \textbf{MT}    & 0.869 $_{0.024}$ & 0.770 $_{0.038}$ & 2.02 $_{0.81}$ & 7.85 $_{2.78}$ & 0.859 $_{0.035}$ & 0.755 $_{0.054}$ & 1.59 $_{0.40}$ & 7.30 $_{2.72}$ \\
    \textbf{UA-MT} & 0.872 $_{0.025}$ & 0.774 $_{0.038}$ & 2.94 $_{2.88}$ & 10.40 $_{10.39}$ & 0.873 $_{0.030}$ & 0.776 $_{0.046}$ & 1.68 $_{0.83}$ & 6.66 $_{3.47}$ \\
    \textbf{BCP}   & \textbf{0.896} $_{0.025}$ & \textbf{0.813} $_{0.040}$ & 1.53 $_{0.31}$ & \textbf{5.87} $_{1.67}$ & 0.901 $_{0.016}$ & 0.821 $_{0.026}$ & 1.34 $_{0.22}$ & 4.71 $_{1.37}$ \\
    \bottomrule
    \end{tabular}
    }
    \label{tab:left_atrium_eval}
    \vspace{-2em}
\end{table}

(3) \textit{Volume vs. Boundary Accuracy}: The results underscore the necessity of boundary metrics. While overlap scores (Dice) can appear tight, surface distances (ASD and HD95) expose severe localization errors. For example, on the 20\% Femur dataset, MT achieves a respectable 0.931 Dice score but exhibits substantially larger HD95 (15.52) compared to BCP (2.05). CAML and BCP consistently yield low boundary errors, while MCF demonstrates the most significant boundary refinement when trained with 40\% labeled data.

These segmentation results establish which methods produce accurate masks, but they do not determine whether the resulting surfaces preserve the population-level shape space needed for SSM analysis.

\begin{table}[!htb]
    \caption{\textbf{Femur Semi-Supervised Segmentation Results.} Metrics include Dice, Jaccard (Jacc.), ASD, and HD95 on test data, reported as mean$_{\text{standard deviation}}$ across nine held-out cases. $\uparrow$/$\downarrow$ indicate higher/lower is better, and the best method per configuration is bolded.}
    \centering
    \setlength{\tabcolsep}{1pt}
    \scalebox{0.86}{
    \begin{tabular}{l|cccc|cccc}
    \toprule
    \multirow{2}{*}{\textbf{Method}} & \multicolumn{4}{c|}{\textbf{20\% Data (8 labeled)}} & \multicolumn{4}{c}{\textbf{40\% Data (16 labeled)}} \\
    \cmidrule(lr){2-5} \cmidrule(lr){6-9}
     & \textbf{Dice} ($\uparrow$) & \textbf{Jacc.} ($\uparrow$) & \textbf{ASD} ($\downarrow$) & \textbf{HD95} ($\downarrow$) & \textbf{Dice} ($\uparrow$) & \textbf{Jacc.} ($\uparrow$) & \textbf{ASD} ($\downarrow$) & \textbf{HD95} ($\downarrow$) \\
    \midrule
    \textbf{DeSCO} & 0.954 $_{0.009}$ & 0.913 $_{0.016}$ & 0.92 $_{0.24}$ & 2.60 $_{0.71}$ & 0.947 $_{0.014}$ & 0.899 $_{0.025}$ & 1.40 $_{0.81}$ & 5.19 $_{4.24}$ \\
    \textbf{SASS}  & 0.950 $_{0.013}$ & 0.905 $_{0.024}$ & 1.58 $_{0.83}$ & 5.14 $_{4.51}$ & 0.958 $_{0.009}$ & 0.920 $_{0.016}$ & 0.77 $_{0.22}$ & 2.48 $_{0.69}$ \\
    \textbf{DTC}   & \textbf{0.961} $_{0.012}$ & \textbf{0.925} $_{0.022}$ & 0.76 $_{0.33}$ & 2.38 $_{1.16}$ & 0.961 $_{0.009}$ & 0.924 $_{0.017}$ & 0.67 $_{0.17}$ & 2.31 $_{0.94}$ \\
    \textbf{MCF}   & 0.959 $_{0.025}$ & 0.922 $_{0.044}$ & 1.57 $_{1.28}$ & 3.99 $_{4.43}$ & \textbf{0.974} $_{0.007}$& \textbf{0.950} $_{0.012}$ & \textbf{0.53} $_{0.21}$ & \textbf{1.49} $_{0.51}$ \\
    \textbf{CAML}  & 0.960 $_{0.010}$ & 0.923 $_{0.012}$ & 0.84 $_{0.22}$ & 2.86 $_{1.15}$ & 0.964 $_{0.006}$ & 0.931 $_{0.011}$ & 0.69 $_{0.12}$ & 2.46 $_{0.58}$ \\
    \textbf{MT}    & 0.931 $_{0.019}$ & 0.871 $_{0.034}$ & 3.99 $_{2.22}$ & 15.52 $_{11.20}$ & 0.923 $_{0.060}$ & 0.863 $_{0.095}$ & 2.32 $_{2.64}$ & 6.70 $_{8.45}$ \\
    \textbf{UA-MT} & 0.944 $_{0.017}$ & 0.894 $_{0.030}$ & 2.73 $_{1.38}$ & 9.67 $_{8.33}$ & 0.946 $_{0.019}$& 0.898 $_{0.034}$ & 1.93 $_{1.93}$ & 7.29 $_{10.28}$ \\
    \textbf{BCP}   & \textbf{0.962} $_{0.004}$ & \textbf{0.926} $_{0.007}$ & \textbf{0.68} $_{0.05}$ & \textbf{2.05} $_{0.09}$ & 0.964 $_{0.016}$ & 0.930 $_{0.026}$ & 0.66 $_{0.22}$ & 2.00 $_{1.37}$ \\
    \bottomrule
    \end{tabular}
    }
    \label{tab:femur_eval}
        \vspace{-2em}
    \end{table}


\subsection{Statistical Shape Modeling Performance Using Different Semi-Supervised Methods}
\begin{figure}[!htb]
    \centering
    \includegraphics[width=\linewidth]{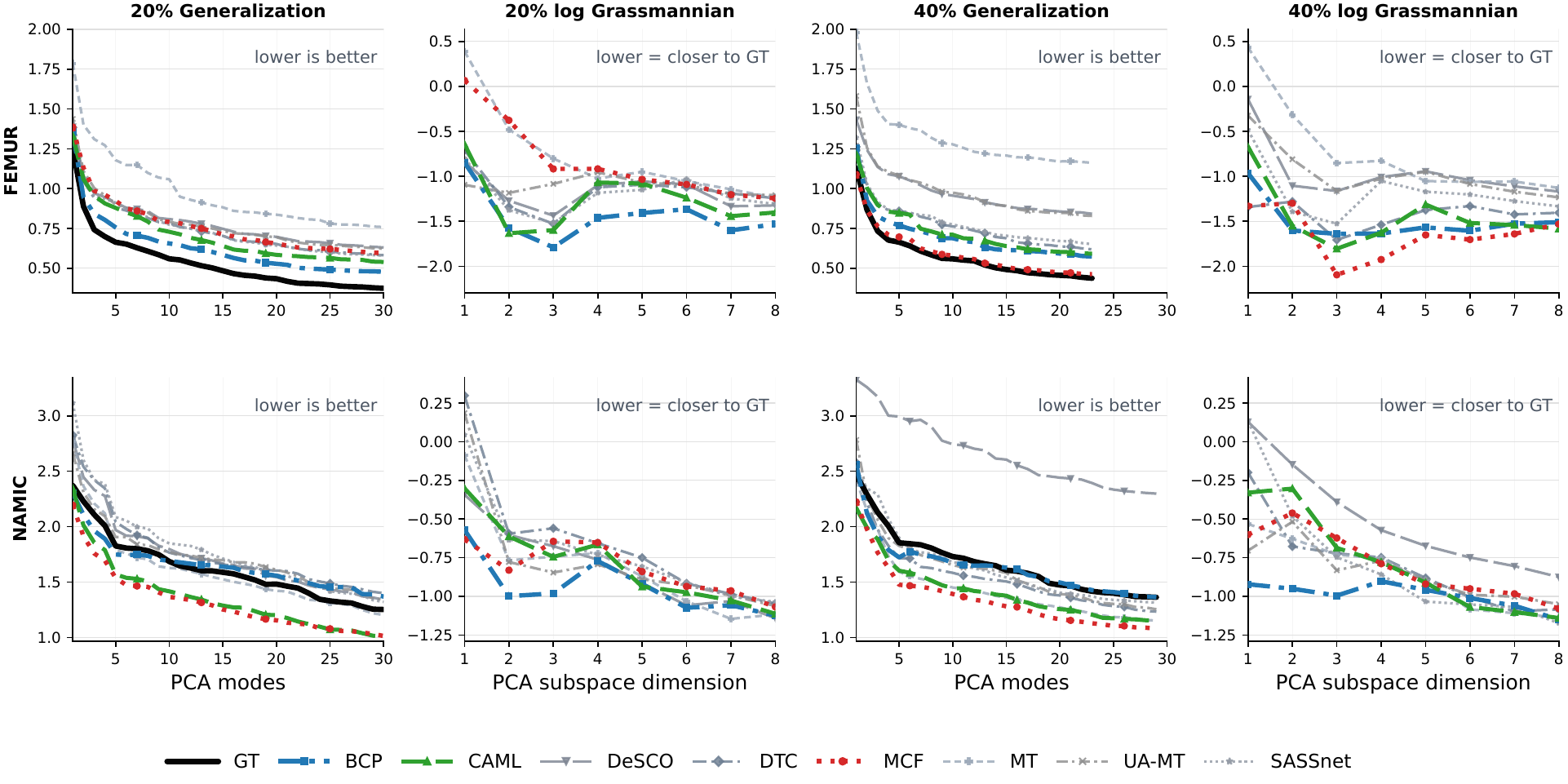}
    \caption{\textbf{Strategy 1: downstream SSM quality when the manual training SSM is available.} Generalization error across PCA modes and log Grassmannian distance to the manual-segmentation SSM subspace are shown for SSMs built from semi-supervised test segmentations and initialized from the manual-segmentation training SSM. Lower values are indicative of better agreement. The generalization curves are indexed by PCA mode, whereas the Grassmannian curves are indexed by PCA subspace dimension; the ranges differ because the latter is limited by the available particle correspondences used to form the subspace.}
    \label{fig:strategy_1_results}
\end{figure}

\textbf{Quantitative results for Strategy 1}. Figure~\ref{fig:strategy_1_results} presents the quantitative evaluation of Strategy 1, where the statistical shape model (SSM) is trained using manually segmented data, while semi-supervised segmentations are used to construct the test SSM. The resulting test SSM is then compared with the corresponding test SSM generated from manual segmentations. Since both approaches use the same training SSM, the evaluation focuses only on the Generalization and Grassmannian distance metrics.

The generalization error measures how closely the SSM constructed from semi-supervised segmentations approximates the ground-truth SSM obtained from manual segmentations. Therefore, a lower generalization error indicates that the semi-supervised segmentations produce shape representations that are more consistent with the ground truth.

The Grassmannian distance evaluates the similarity between the PCA subspaces of the manually generated and semi-supervised test SSMs. As a dimensionless metric, smaller values indicate greater similarity between the learned shape spaces. However, while a low Grassmannian distance suggests that the underlying PCA subspaces are closely aligned, it does not necessarily imply that the two models will exhibit equivalent shape reconstruction or generalization performance.

For the FEMUR dataset, test SSMs constructed from manual segmentations (GT) consistently achieve lower generalization errors across all PCA modes than those constructed from semi-supervised predictions, irrespective of whether 20\% or 40\% of the training data is annotated. Among the semi-supervised methods trained with 20\% annotations, BCP most closely matches the GT model in terms of generalization performance, followed by CAML. The Grassmannian distance analysis shows that BCP also produces the closest PCA subspace to the GT-based SSM, with CAML ranking second. When 40\% of the training data is annotated, MCF improves its generalization performance and becomes the second-best method after GT, while BCP ranks third. Nevertheless, BCP continues to exhibit the lowest Grassmannian distance, with CAML and MCF closely following. Furthermore, the overall Grassmannian distances decrease when using 40\% annotated data, indicating that increasing annotation availability improves the agreement between the PCA subspaces learned from semi-supervised and manually segmented test SSMs.

The NAMIC dataset exhibits a different trend. Several semi-supervised methods achieve lower generalization errors than the GT-based SSM, particularly when trained with 20\% annotated data, and this behavior persists for most methods when 40\% of the data is annotated. However, these improvements in generalization do not imply that the learned statistical shape models recover the same population-level shape statistics as the manual-segmentation reference. The Grassmannian distance remains comparable to that observed for the FEMUR dataset, indicating that although the semi-supervised SSMs reconstruct test shapes effectively, their PCA subspaces remain distinct from the GT SSM.

Generalization measures how accurately an SSM reconstructs unseen test correspondences, whereas the Grassmannian distance quantifies the similarity between the PCA subspaces of two SSMs. Consequently, a lower generalization error does not necessarily indicate recovery of the same anatomical modes of variation. Instead, the improved generalization observed for some semi-supervised methods may result from smoother or more internally consistent predicted segmentations, reduced annotation variability, or differences in the distributions of the training and test surfaces. Such behavior is expected for many semi-supervised approaches that incorporate entropy minimization or consistency regularization, which encourage consistent predictions across the dataset while potentially altering the underlying statistical shape basis.

\begin{figure}[!htb]
    \centering
    \includegraphics[trim={0.25cm 0.35cm 0.25cm 0.0cm}, clip=true, width=0.85\linewidth]{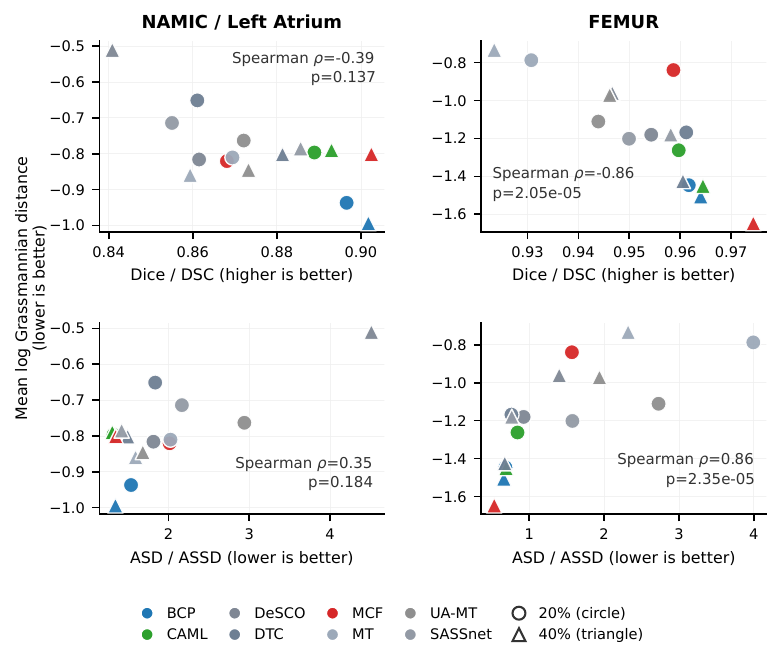}
    \caption{\textbf{Segmentation accuracy versus downstream SSM subspace agreement.}
    Each point represents a semi-supervised method at one annotation setting, with circles for 20\% and triangles for 40\% labeled data. Dice/DSC and ASD/ASSD are plotted against the mean log Grassmannian distance from Strategy~1; higher Dice is better, while lower ASD and Grassmannian distance indicate better boundary accuracy and closer agreement with the manual SSM subspace.}
    \label{fig:seg_vs_grassmannian}
    \vspace{-2em}
\end{figure}

To evaluate whether segmentation accuracy predicts downstream SSM quality, Fig.~\ref{fig:seg_vs_grassmannian} compares Dice and ASD with Strategy~1 Grassmannian distance. Strong correlations can be observed for the femur but not for the left atrium, indicating that segmentation accuracy is not a consistent predictor of downstream shape-space quality across anatomies and that the relationship is anatomy-dependent.

\begin{figure}[!htb]
    \centering
    \includegraphics[width=0.85\linewidth]{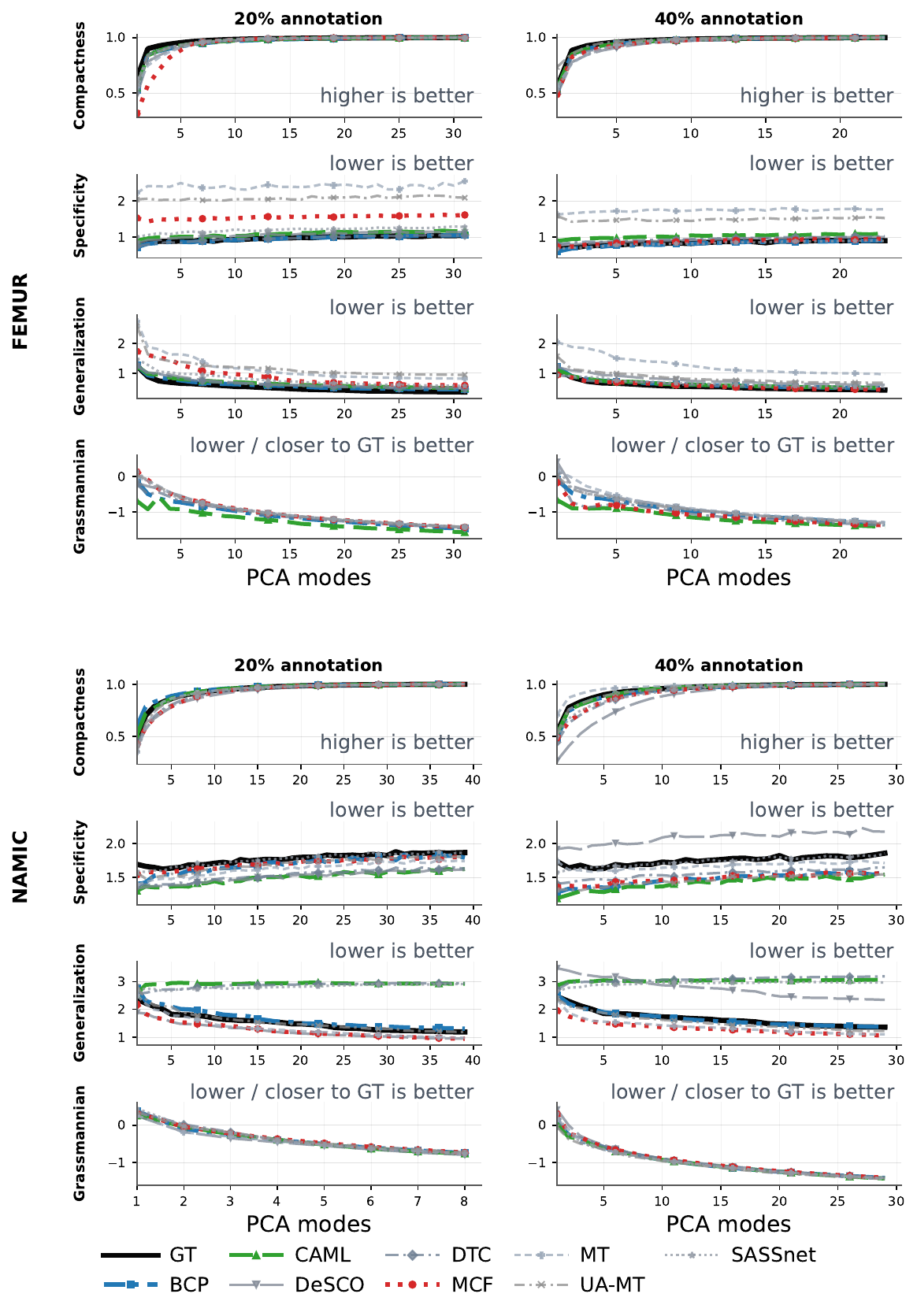}
    \caption{\textbf{Strategy 2: downstream SSM quality when both training and test SSMs are built from semi-supervised segmentations.} Compactness, specificity, generalization, and log Grassmannian distance are shown for FEMUR and NAMIC under 20\% and 40\% labeled-data settings. Higher compactness indicates a more efficient shape model; lower specificity, generalization, and Grassmannian distance indicate better shape fidelity or closer agreement with the manual-segmentation reference.}
    \label{fig:strategy_2_results_20_40_percent}
    \vspace{-2em}
\end{figure}

\textbf{Quantitative Results for Strategy 2.} Figure~\ref{fig:strategy_2_results_20_40_percent} evaluates Strategy 2, the stricter deployment setting in which both training and test SSMs are constructed from semi-supervised predictions. It reports compactness, specificity, generalization and Grassmannian distance for FEMUR and NAMIC datasets for models trained with 20\% and 40\% of annotations. 

For FEMUR, the ground truth SSM achieves the highest AUC for compactness, indicating that it is the most compact model, followed closely by CAML and BCP for both 20\% and 40\% annotated models. Generalization scores show a similar trend, with the GT SSM consistently performing best, followed by BCP and DTC as the second and third best models for 20\% annotation, while MCF and DTC rank second and third for 40\% annotation. In terms of specificity, DTC closely matches the GT SSM, with DeSCO ranking third for 20\% models. For 40\% models, SASSnet and BCP produce specificity scores that are also very similar to the GT SSM. The Grassmannian distance remains comparable to that observed in FEMUR results from Strategy 1, indicating that the PCA subspace of test SSMs—whether initialized with the GT train SSM or the predicted train SSM—remains similarly distant.

For NAMIC, results align with Strategy 1, where the GT model does not achieve the best generalization performance. Instead, DeSCO, MCF, and DTC emerge as the top-performing models in generalization. The GT model is also not the most compact; BCP performs best for 20\% annotated models, while DTC ranks highest for 40\% models. A similar pattern is observed for specificity scores. Unlike FEMUR, where increasing annotated data has little impact—as reflected by the unchanged Grassmannian distance—NAMIC shows a slight improvement in Grassmannian distance when moving from 20\% to 40\% annotation, suggesting that additional annotations provide a small but positive effect on SSM subspace alignment.



\subsection{Qualitative Analysis}
\begin{figure}[!htb]
    \centering
    \includegraphics[width=\linewidth]{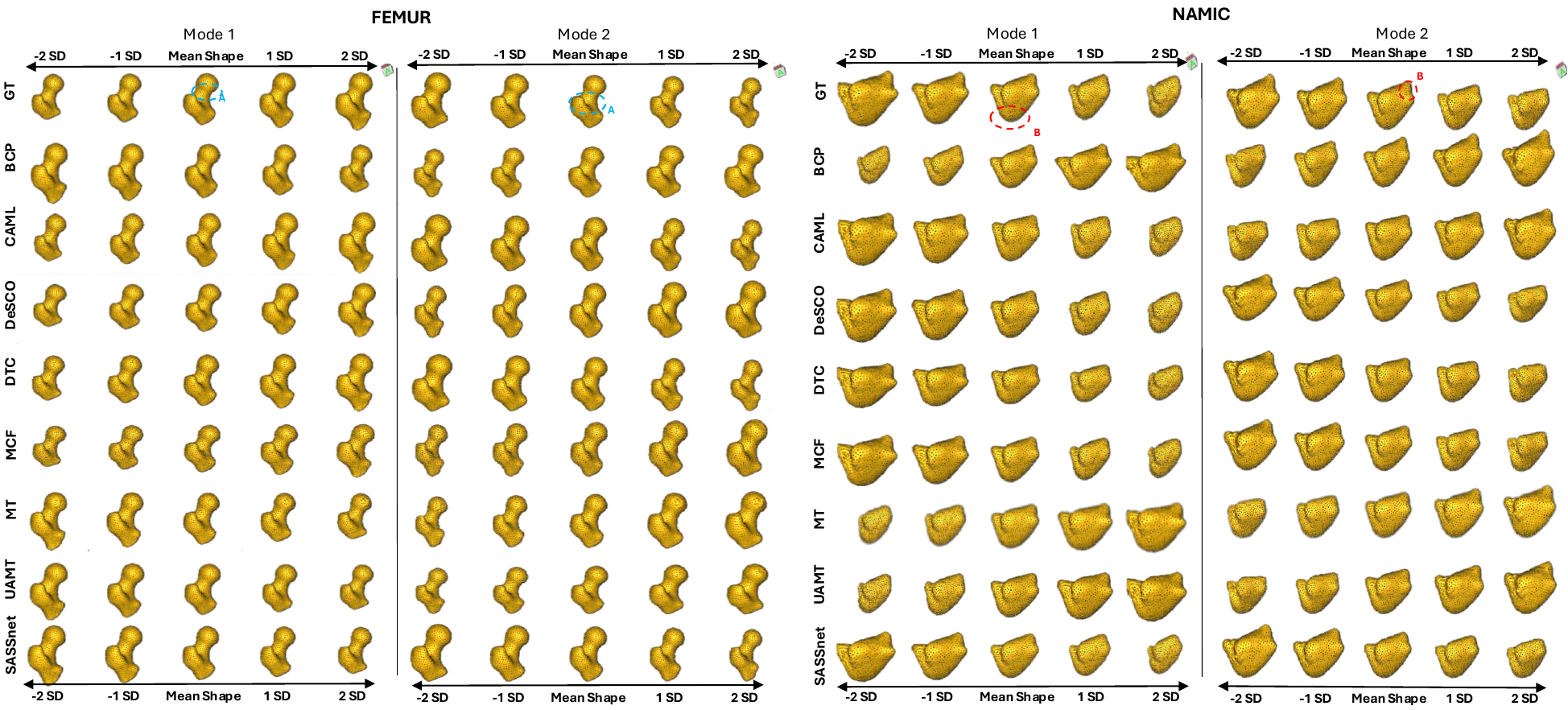}
     \caption{\textbf{Strategy 1 qualitative mode analysis.} Representative first and second modes of variation for FEMUR and left atrium SSMs, using the manual training SSM and test SSMs from semi-supervised segmentations with 20\% labeled data. Each row shows shapes at $-2\sigma$, $-1\sigma$, mean, $+1\sigma$, and $+2\sigma$. Annotation A highlights femoral variation near the head--neck region, while annotation B highlights thin left-atrial structures near the pulmonary vein/ostial region. Highlighted regions are illustrative and do not imply other regions are unchanged.}
    \label{fig:setting1_qualitative}
\end{figure}

\begin{figure}[!htb]
    \centering
    \includegraphics[width=\linewidth]{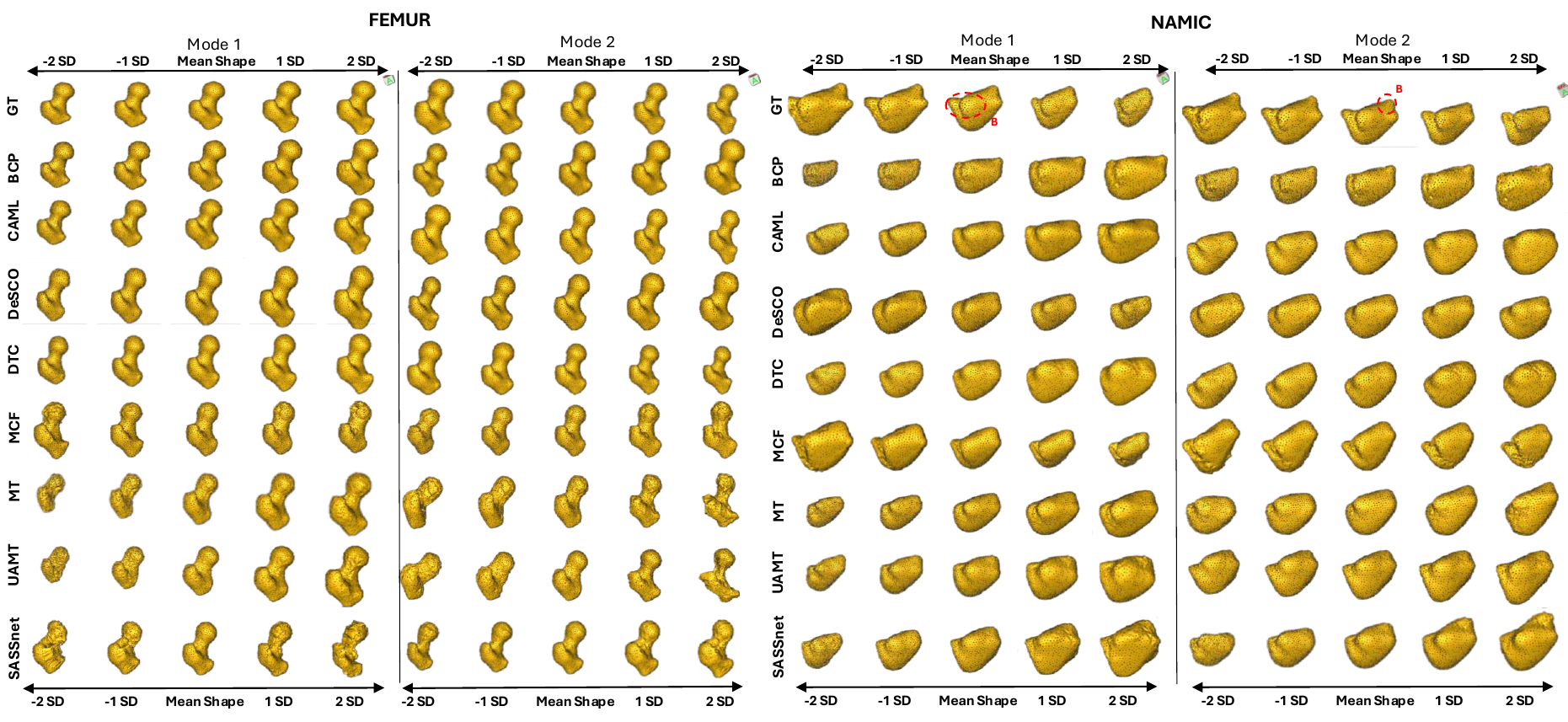}
    \caption{\textbf{Strategy 2 qualitative mode analysis.} Representative first and second modes of variation for FEMUR and left atrium SSMs, with both training and test SSMs constructed from semi-supervised segmentations using 20\% labeled data. Each row shows shapes at $-2\sigma$, $-1\sigma$, mean, $+1\sigma$, and $+2\sigma$. Annotation B highlights thin left-atrial structures near the pulmonary vein/ostial region, where differences in smoothness and anatomical detail are most visible. Highlighted regions are illustrative and do not imply other regions are unchanged.}
    \label{fig:setting2_qualitative}
    \vspace{-1em}
\end{figure}

Figures~\ref{fig:setting1_qualitative} and~\ref{fig:setting2_qualitative} provide representative qualitative examples of the first two modes of variation under the most annotation-limited setting. We show the 20\% setting because it is the most annotation-constrained method in our setting and therefore most clearly exposes shape-model failure modes.

\textbf{Qualitative results for Strategy 1.} Figure~\ref{fig:setting1_qualitative} illustrates representative first and second modes of variation for Strategy 1. The results demonstrate that nearly all methods generate smooth and interpretable statistical shape models (SSMs), effectively capturing both the first and second modes. However, in some cases, the direction of variation is reversed in the test SSMs produced by semi-supervised models. Specifically, BCP, MT, and UA-MT exhibit this reversal for both NAMIC and FEMUR in modes 1 and 2, while SASSnet also shows a reversed mode of variation for NAMIC in mode 1. Despite these reversals, most models generate qualitatively smooth, meaningful, and interpretable representations of dominant shape variation.

\textbf{Qualitative Results for Strategy 2.} Figure~\ref{fig:setting2_qualitative} illustrates representative first and second modes of variation for Strategy 2. From these figures, we observe that SSMs generated for the FEMUR dataset by BCP, CAML, DeSCO, and DTC exhibit smooth shapes, closely resembling the ground truth SSM, while other methods result in poor-quality SSMs. Additionally, CAML, DeSCO, and DTC effectively capture the first mode of variation, though are noticeable differences in BCP’s results. A similar trend is observed for NAMIC, where the same semi-supervised methods also produce smooth SSMs. However, none of the SSM models successfully capture the pulmonary vein ostia (left atrial protrusions), as the produced SSMs exhibit overly smooth and poorly defined anatomical structures, leading to significantly noticeable discrepancies in both mode 1 and mode 2 compared to the GT SSM.

\textbf{Practical guidelines for Usage of Semi-supervised methods for SSMs}. Segmentation metrics alone were not sufficient to predict downstream SSM quality. Although lower ASD and HD95 generally correlated with improved shape modeling, methods with the lowest boundary errors did not always produce the most accurate SSMs. Likewise, models with competitive Dice scores or lower generalization error did not necessarily recover the PCA subspace derived from manual segmentations. Methods incorporating boundary refinement and pseudo-label correction, such as BCP, MCF, and CAML, more consistently preserved shape-space quality, but no single method performed best across all anatomies and annotation budgets.

Qualitative analysis further showed that, despite achieving superior segmentation accuracy, methods such as CAML, BCP, and MCF did not always reproduce the same principal modes of variation as models built from manual segmentations. This suggests that optimizing for segmentation consistency alone may not be sufficient to preserve the true anatomical variability of a population. A promising direction for future semi-supervised methods is to explicitly model population-level shape variation, for example through statistical shape priors, distribution-aware learning, or conditioning segmentation on anatomical and image-texture variability. Such approaches may better preserve the underlying distribution of organ shapes while maintaining segmentation accuracy.

Overall, these findings highlight the importance of validating semi-supervised segmentation using downstream shape-space metrics before applying it to clinical morphology studies or shape-based biomarkers. They also suggest that semi-supervised segmentations may be more suitable for constructing test SSMs than training SSMs. While test models primarily evaluate new shapes against an existing statistical model, training SSMs must accurately capture the full distribution of anatomical variability. The observed loss of agreement in the principal modes of variation indicates that current semi-supervised methods may not faithfully preserve this true variability, potentially limiting their suitability for building population-representative training SSMs.

\textbf{Limitations}. This study evaluated two anatomies using default implementations of representative semi-supervised methods. Results may differ with architecture tuning, alternative annotation protocols, or anatomies with more complex topology. Because accurate shape-space reconstruction is a prerequisite for morphology-based analysis, this work focused on SSM quality rather than clinical prediction. The datasets are modest in size, with 49 FEMUR volumes and 59 NAMIC volumes, which limits the statistical power of method comparisons and motivates validation on larger sample sizes as part of future work which will also investigate whether differences in PCA subspace alignment translate to improved disease classification, longitudinal progression modeling, and other clinical applications.

\section{Conclusion}
We presented a benchmark for evaluating whether semi-supervised segmentation methods can replace manual segmentations for statistical shape modeling. Our results show that annotation-efficient segmentation can substantially reduce manual labeling requirements, but downstream SSM quality depends strongly on the semi-supervised method, anatomy, and evaluation metric. In particular, generalization and segmentation overlap metrics can obscure changes in the learned shape subspace, motivating the use of Grassmannian subspace analysis and qualitative comparison of PCA modes of variation as complementary measures of SSM quality. These findings provide practical guidance for deploying semi-supervised segmentation in shape-driven population studies and suggest that current semi-supervised methods, while effective at improving segmentation consistency, do not necessarily preserve the underlying population-level anatomical variability. Future semi-supervised approaches should therefore incorporate mechanisms that explicitly preserve shape distributions and population statistics, enabling segmentations that more faithfully reflect true anatomical variation while maintaining high segmentation accuracy.

{\fontsize{9pt}{11pt}\selectfont 
\textbf{Disclosure of Interests.}
The authors have no competing interests to declare that are relevant to the content of this article.
\par}

{\fontsize{9pt}{11pt}\selectfont
\textbf{Acknowledgments}.The National Institutes of Health supported this work under grant numbers NIBIB-U24EB029011 and NIAMS-R01AR076120. The content is solely the authors’ responsibility and does not necessarily represent the official views of the National Institutes of Health.
\par}


\newpage
\bibliographystyle{splncs04}
\bibliography{bib}
\end{document}